\documentclass[conference]{IEEEtran}
\IEEEoverridecommandlockouts

\pdfoutput=0
\usepackage{amsmath,amssymb,amsfonts}
\usepackage{textcomp}
\def\BibTeX{{\rm B\kern-.05em{\sc i\kern-.025em b}\kern-.08em
		T\kern-.1667em\lower.7ex\hbox{E}\kern-.125emX}}
\DeclareMathOperator*{\argmax}{arg\,max}

\DeclareMathOperator{\Var}{Var}

\DeclareMathOperator{\E}{\mathbb{E}}
\newcommand{\subparagraph}{}

\usepackage{graphicx}
\usepackage{epstopdf}
\usepackage{color}
\usepackage[explicit]{titlesec}
\usepackage{textcomp}
\usepackage{multirow}
\usepackage{ifpdf}

\usepackage[letterpaper, left=0.625in, right=0.625in, bottom=0.5in, top=0.484in]{geometry}
 \usepackage[caption=false,font=footnotesize]{subfig}
\def\BibTeX{{\rm B\kern-.05em{\sc i\kern-.025em b}\kern-.08em
    T\kern-.1667em\lower.7ex\hbox{E}\kern-.125emX}}

\titlespacing{\subsection}{0pt}{4pt}{2pt}
\color{black}

\title{Experimental Analysis of Reinforcement Learning Techniques for Spectrum Sharing Radar 
	{\footnotesize \textsuperscript{}}
	\thanks{Accepted for Publication at 2020 IEEE International Radar Conference, to appear April 2020. This is the authors' version of the work. \newline $^\dag$C.E. Thornton and R.M. Buehrer are with Wireless@VT, Bradley Department of ECE, Virginia Tech, Blacksburg, VA, 24061, USA. \newline \indent $\ddagger$A.F. Martone and K.D. Sherbondy are with the US Army Research Laboratory, Adelphi MD, 20783, USA. \newline \indent The support of US Army Research Office (ARO) grant W911NF-15-2-0066 is gratefully acknowledged.}}

\author{\IEEEauthorblockN{Charles E. Thornton$^{\dag}$, R. Michael Buehrer$^{\dag}$, Anthony F. Martone$^{\ddagger}$, and Kelly D. Sherbondy$^{\ddagger}$}}

\begin{document}


\maketitle

\begin{abstract}	
In this work, we first describe a framework for the application of Reinforcement Learning (RL) control to a radar system that operates in a congested spectral setting. We then compare the utility of several RL algorithms through a discussion of experiments performed on Commercial off-the-shelf (COTS) hardware. Each RL technique is evaluated in terms of convergence, radar detection performance achieved in a congested spectral environment, and the ability to share 100MHz spectrum with an uncooperative communications system. We examine policy iteration, which solves an environment posed as a Markov Decision Process (MDP) by directly solving for a stochastic mapping between environmental states and radar waveforms, as well as Deep RL techniques, which utilize a form of \textit{Q}-Learning to approximate a parameterized function that is used by the radar to select optimal actions. We show that RL techniques are beneficial over a Sense-and-Avoid (SAA) scheme and discuss the conditions under which each approach is most effective.
\end{abstract}
\vspace{2mm}
\begin{IEEEkeywords}
cognitive radar, spectrum sharing, Markov decision process, deep reinforcement learning, radar detection
\end{IEEEkeywords}

\section{Introduction}
The Third Generation Partnership Project (3GPP) has recently received FCC approval to support 5G New Radio (NR) operation in sub-6 GHz frequency bands that are heavily utilized by radar systems \cite{b1,fcc}. Thus, there is a significant need for radar systems capable of dynamic spectrum sharing. Since radars consume large amounts of spectrum in many of the sub-6 GHz frequencies being repurposed for 5G NR use, interest has grown in cognitive algorithms that allow radar and communication systems to share increasingly congested spectrum with minimal mutual interference \cite{future,haykin}. As federal spectrum policies and 3GPP standards continue to implement spectrum sharing, next-generation radar systems must have the ability to quickly sense open spectrum and adapt intelligently during short windows of opportunity. 

One proposed method for adaptive radar waveform selection is Sense-and-Avoid (SAA). SAA identifies occupied portions of the spectrum and directs radar transmissions to the largest contiguous open bandwidth \cite{ben}. However, SAA does not learn to recognize patterns, which can be used to avoid Radio Frequency Interference (RFI) in a strategic manner, where some RFI is avoided and interference with other signals may be allowed depending on application-specific preferences. Thus, the performance of SAA is fixed based on the environment, and can be ineffective when the interference channel changes rapidly. Additionally, a fully adaptive radar framework based on minimization of a tracking cost function with multiple objectives has been proposed \cite{cost}. However, this procedure does not consider coexistence with other RF emitters, such as communication systems. Here, we demonstrate the effectiveness of machine learning techniques that estimate optimal radar transmission strategies based on a model of the radar's environment as a Markov Decision Process (MDP). Experimental detection characterization is a performed using a software defined radar (SDRadar) prototype. The concept of applying Reinforcement learning (RL) to control a radar waveform selection was introduced and simulated in \cite{selvi}. However, the simulations used for verification only examined the radar's performance in the presence of a single simulated emitter and the system model suffers from the curse of dimensionality. The use of Deep RL to reduce the dimensionality of the learning problem was introduced in \cite{b12}, also in a simplified simulation-based setting. Here, we develop a general RL procedure for determining long-term radar behavior in a congested RF environment. Both iterative and Deep RL techniques are discussed, and this framework can be used extended to more complex RL architectures. The effectiveness of the RL techniques is evaluated on a SDRadar platform implemented on a Universal Software Radio Peripheral (USRP). We demonstrate that our radar system is able to identify the presence of recorded RFI and estimate the optimal behavior for reduced mutual interference and increased target detection performance in a congested environment.

A major challenge for frequency agile cognitive radar systems is target distortion in the range-Doppler processed data as a result of clutter modulation from intra-CPI pulse adaptations \cite{b11}. RL-based radar control is useful to mitigate this problem since radar behavior is motivated through a human-defined reward mapping. For example, the radar operator could introduce a penalty for intra-CPI radar adaptations, or introduce a reward for using the same frequency bands without interference over a long period of time. Here we demonstrate that using RL techniques motived by a reward function that balances SINR and bandwidth utilization, along with a penalty on excessive intra-CPI pulse adaptations results in desirable radar operation from both a target detection and spectrum sharing perspective. 

\subsection{Contributions}

The novelty of this work lies in the first detailed comparison of multiple RL techniques for the radar waveform selection process. Our comparison experimentally analyzes each algorithm in terms of convergence properties, radar performance characteristics, and spectrum sharing capabilities. This paper also introduces adaptive radar waveform selection based on Deep Recurrent Q-Learning (DRQL) and Double Deep Q-Learning (DDQL) for the first time. Further, we show our approaches outperform a basic SAA approach in realistic coexistence scenarios.

The remainder of this paper is structured as follows. In Section II, we discuss our MDP formulation and the RL framework developed to control the agent. In Section III, the SDRadar prototype system used for testing is described and our experimental procedure is outlined. In Section IV, results for convergence of learning, radar performance improvement, and spectrum sharing capabilities are discussed.  

\section{MDP Formulation and Radar System Model}

\begin{figure}[t]
	\centering
	\includegraphics[scale=0.28]{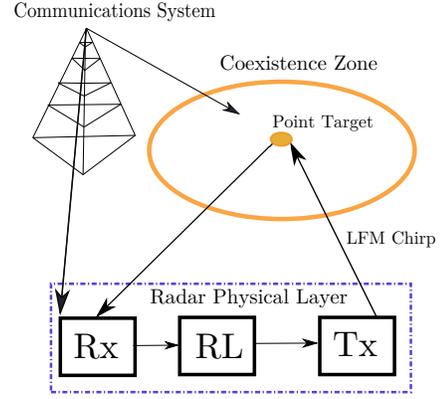}
	\caption{Model of the radar/cellular coexistence scheme. The radar sends and receives LFM chirp waveforms. The bandwidth and center frequency of the radar's waveform are modified based on behavior learned from the reinforcement learning process. }
	\label{fig:model}
\end{figure}

The cognitive radar system described here models the radar waveform selection process as a MDP, which is a mathematical model for human decision making \cite{rl}. A visualization of the radar environment, which consists of the cognitive radar system, a point target, and a communications system, can be seen in Figure \ref{fig:model}. The idea of using a MDP model for a radar environment is introduced in detail in \cite{selvi}. Here, we briefly describe our system's MDP environmental model and proceed to a discussion of RL algorithms, which learn a mapping between environmental states and radar transmissions to optimize the radar's operation in a given environment.

\indent A MDP is specified by the tuple $[S, A, \Gamma, R, \gamma, \pi^*]$. The state space $S = \{s_{1},s_{2},...s_{n}\}$ is the complete set of possible environmental states the radar may experience. The action space $A = \{a_{1},a_{2},...a_{n}\}$ contains the set of all actions that the radar may take to transition between states. The transition probability function $\Gamma(s,a,s')$ denotes the probability that the agent transitions from state $s$ to $s'$ while taking action a. Since spectrum observations do not tell us this information in advance, we estimate $\Gamma$ using a frequentist approach.  
\\
\indent Here, the state space consists of 100MHz spectrum split into five sub bands. At each time step $t$, the interference state $s_{t}$ is expressed as a vector of binary values where zero designates an open channel and one denotes a channel occupied by the communications system. For example a state of $s_{t} = [0, 0, 0, 1, 1]$ corresponds to interference power over threshold $P_{0}$, while the observed power in the left three bands is under $P_{0}$ and are considered available for radar operation. The available actions consist of transmitting a linear frequency modulated (LFM) chirp waveform in any set of contiguous sub bands. In this case, the optimal action is to transmit in the left three bands, resulting in an action vector of $a_{t} = [1, 1, 1, 0, 0]$. While it would be desirable to split the spectrum into more than five sub bands, and therefore have more possible actions, this would exponentially increase the state space, which presents a dimensionality problem for the learning process. The total number of valid actions, $N_{A}$, given $N$ sub bands, is 
\begin{equation}
N_{A} = \textstyle{\sum_{i=0}^{N}} i = \frac{N(N+1)}{2},
\end{equation}

\noindent while the total number of interference states, $N_{S}$ for $N$ sub-bands, where the past $M$ states are considered is
\begin{equation}
N_{S} = 2^{M*N}.
\end{equation}

\indent The reward function is represented as $R(s,a,s')$ and defines the reward the agent receives from transitioning from state $s$ to $s'$ by taking action $a$. The MDP is characterized by the unique choice of transition and reward functions. These rewards can take any numerical value and can correspond to a variety of results, allowing the system operator to define ideal behavior based on the desired application. Here, we choose primarily to balance a fundamental trade-off between bandwidth, SINR, and pulse adaptation. We highly value SINR from a detection standpoint, but also require that the radar use enough bandwidth to achieve sufficient range resolution for separation of nearby targets and tracking applications, as range resolution $\Delta R \propto 1/Bandwidth$. While both SINR and bandwidth are desired, the radar must also limit intra-CPI adaptations to avoid target distortion. Rewards are used to leverage domain knowledge and thus can be modified to accommodate the specific RF scenario or radar application.

The discount factor is represented as $\gamma$ $\in$ [0,1] and defines the agent's emphasis on long term rewards. When $\gamma$ is close to 1, the agent demonstrates a strong preference for long term rewards, and conversely prefers immediate rewards as $\gamma \rightarrow 0$. the MDP has been specified, we describe methods for finding solutions. 
\subsection{Policy Iteration}

The first technique for solving the MDP discussed here is the policy iteration algorithm, which involves solving the MDP explicitly to find a policy, $\pi$, which maps states to actions. First, we define the well-known value function, $V^{\pi}(s)$ that determines the agent's preference for state $s$ while following policy $\pi$
\begin{equation} \begin{aligned} 
V^\pi(s) = \mathbb{E} \bigg[\textstyle{\sum_{k = 0}^{\infty}} \gamma^k R_{t+k} \bigg| \pi, s_t = s \bigg],
\end{aligned} \end{equation}

\noindent where $\mathbb{E}[x]$ is the expectation of $x$ and $R_{t+k}$ is the reward obtained at time $t+k$.

\indent The optimal policy, $\pi^*(s)$, is found by the value function $V^{\pi^{*}}(s)$ where $V^{\pi^{*}}(s)  \geq V^{\pi}(s) \; \forall_{\pi,s}$. This ensures that the agent receives the greatest expected reward for the given environment. The optimal policy $\pi^*$ is then
\begin{equation} \begin{aligned} 
	 \pi^{*}(s)= \argmax_{\pi} \; \mathbb{E} \bigg[ \textstyle{\sum_{t=0}^{\infty}} \gamma^t R(s_t) \bigg| \pi \bigg],  
	 \end{aligned}	
\end{equation}

\noindent where $R(s_{t})$ is the reward obtained from being in state $s_{t}$. \\
\indent To solve for $\pi^{*}$ we first update the policy $\pi$ to maximize the expected utility of the subsequent state $s'$, which yields an updated policy $\pi'(s)$
\begin{equation}
\pi'(s) = \argmax_{a \in A} \; \big[ \textstyle{\sum_{s' \in S}} \; \Gamma(s,a,s') V^{\pi}(s') \big].
\end{equation}

\indent The updated value function $V^{\pi'}(s)$ is then computed using (3) and the process repeats until the policy does not change and we reach stable solution $\pi^{*}$. This technique performs all learning during an offline training phase where the policy is developed. The radar then acts on the learned policy until it is re-trained.
\subsection{Deep RL Methods}

To reduce the complexity of the RL problem, which requires large, sparse, transition and reward matrices in the policy iteration approach, we can also use Deep Learning techniques, which solve the underlying MDP via function approximation without explicitly solving (2). Here we base our approaches on the DQL algorithm, first described in \cite{b15}. Instead of computing $\pi^{*}$ directly, which requires inversions of the reward and transition matrices, we estimate the \textit{quality function}, $Q^{\pi}(s,a)$, which is similar to $V^{\pi}(s)$ above and can be written as
\begin{equation}
Q^{\pi}(s,a) = \E[R_{t} | s = s_{t},a = a_{t},\pi].
\end{equation}

\indent The goal of DQL is to directly approximate $Q^{*}(s,a)$, the function which maximizes the expected reward for all observed states. This is equivalent to taking the supremum over all possible $\pi$
\begin{equation}
Q^{*}(s,a) = \sup_{\pi} Q^{\pi}(s,a).
\end{equation}
\indent The estimation of $Q^{*}(s,a)$ is performed by training a deep neural network with Stochastic Gradient Descent (SGD) to update the network weights. Since function approximation in RL has traditionally yielded unstable or divergent results when a non-linear function approximator is used, the DQL approach uses two neural networks to stabilize the learning process. The additional network is known as a \textit{target network}, which remains frozen for a large number of time steps and is updated gradually based on the estimation from a separate neural network which is updated on a much faster time scale. This allows for smooth function approximation in the presence of noisy measurements. The target values can be written as
\begin{equation}
y = r + \gamma \argmax_{a'} Q(s',a';\theta^{-}_{i}),
\end{equation}
\noindent where $\theta^{-}_{i}$ corresponds to the frozen target network's weights from the most recent update and $r = R(s,a,s')$. The loss function that we seek to minimize then varies over each iteration $i$ and is expressed as
\begin{equation*}
L_{i}(\theta_{i}) = \E_{s,a,r,s'} \left[ (\E_{s'}[y|s,a]-Q(s,a;\theta_{i})^2) \right]
\end{equation*}
\begin{equation}
= \E_{s,a,r,s'}[(y-Q(s,a;\theta_{i}^2)] + \E_{s,a,r,s'}[\Var_{s'}[y]],
\label{loss}
\end{equation}

\indent where the last term in Equation \ref{loss} is the variance of the target network weights. Since the target network weights, $\theta'_{i}$, are frozen, they do not depend on the weights we are currently optimizing, $\theta_{i}$. Thus, this term can be ignored in practice. However, the loss function above involves taking an expectation over all $(s,a,r,s')$ samples, which is impractical for real applications. To approximate this expectation, we use a technique called \textit{experience replay}, which involves storing observed transitions, $\phi(s,a,r,s')$, in a memory bank, $\chi(\phi),$ and sampling random $\phi \in \chi(\phi)$ with the goal of obtaining uncorrelated samples. The sampled transitions are then used to perform SGD updates. Since the distribution of radar behavior is averaged over many previous states, this technique allows for smooth function approximation and avoids divergence in the parameters \cite{b15}

\indent Using the sampled transitions, we arrive at the gradient update given by
\begin{equation*}
\theta_{s,a} = \theta_{s,a} - \alpha \frac{\partial L(\theta) }{\partial \theta_{s,a}}
\end{equation*}
\begin{equation}
 = (1- \alpha)\theta_{s,a} + \alpha(R(s,a,s')) + \gamma* \argmax_{a'}Q(s',a';\theta))
\end{equation}

\noindent where $\alpha$ is the learning rate, or step size, of the SGD algorithm. 

\indent The radar RL process consists of two phases, \textit{offline learning} and \textit{online learning}. During the offline learning phase the radar simulates random actions drawn from a uniform distribution over $A$. This phase allows the radar to explore the space of possible actions and update the network weights accordingly. However, the radar does not send waveforms during this phase. After the offline learning phase, the radar enters the online learning phase, during which the radar sends waveforms based on the $Q^{*}(s,a)$ approximated during the offline learning phase. In addition to normal radar operation, the associated $R(s,a,s')$ is calculated for every transmission and the network weights are updated via SGD after a fixed number of transmissions. While the radar is still able to learn during this phase, its exploration of the action space is more limited, since the current actions are guided by behavior learned during the offline phase.

\indent In the RL literature, many extensions to the basic DQL algorithm have been proposed \cite{rlsurvey}. Here, we focus our discussion on two of the most prominent examples, Double Deep $Q$-Learning (DDQL) and Deep Recurrent $Q$-Learning (DRQL). The DDQL algorithm was proposed to mitigate upward bias in the approximation of $Q^{*}(s,a)$, which can be caused by estimation errors of any kind \cite{ddqn}. Estimation errors are inherent to large-scale learning problems and can be caused by environmental noise, non-stationarity, function approximation, and many other sources. DDQL minimizes upward bias by attempting to decouple action selection and action evaluation. This is done by evaluating an action using the online network, as in DQL, but using the target network to estimate the value of the the action. The target for DDQL is written as
\begin{equation}
Y_{Double} = r + \gamma Q(s, \;\argmax_{a}Q(s,a;\theta); \;\theta^{-}),
\end{equation}

\noindent while the target network update remains the same as in the case of DQL. This basic extension to DQL has been shown to reduce overoptimism when tested relative to DQL on deterministic video games.

\begin{figure*}[t]
	\centering
	\includegraphics[scale=0.5]{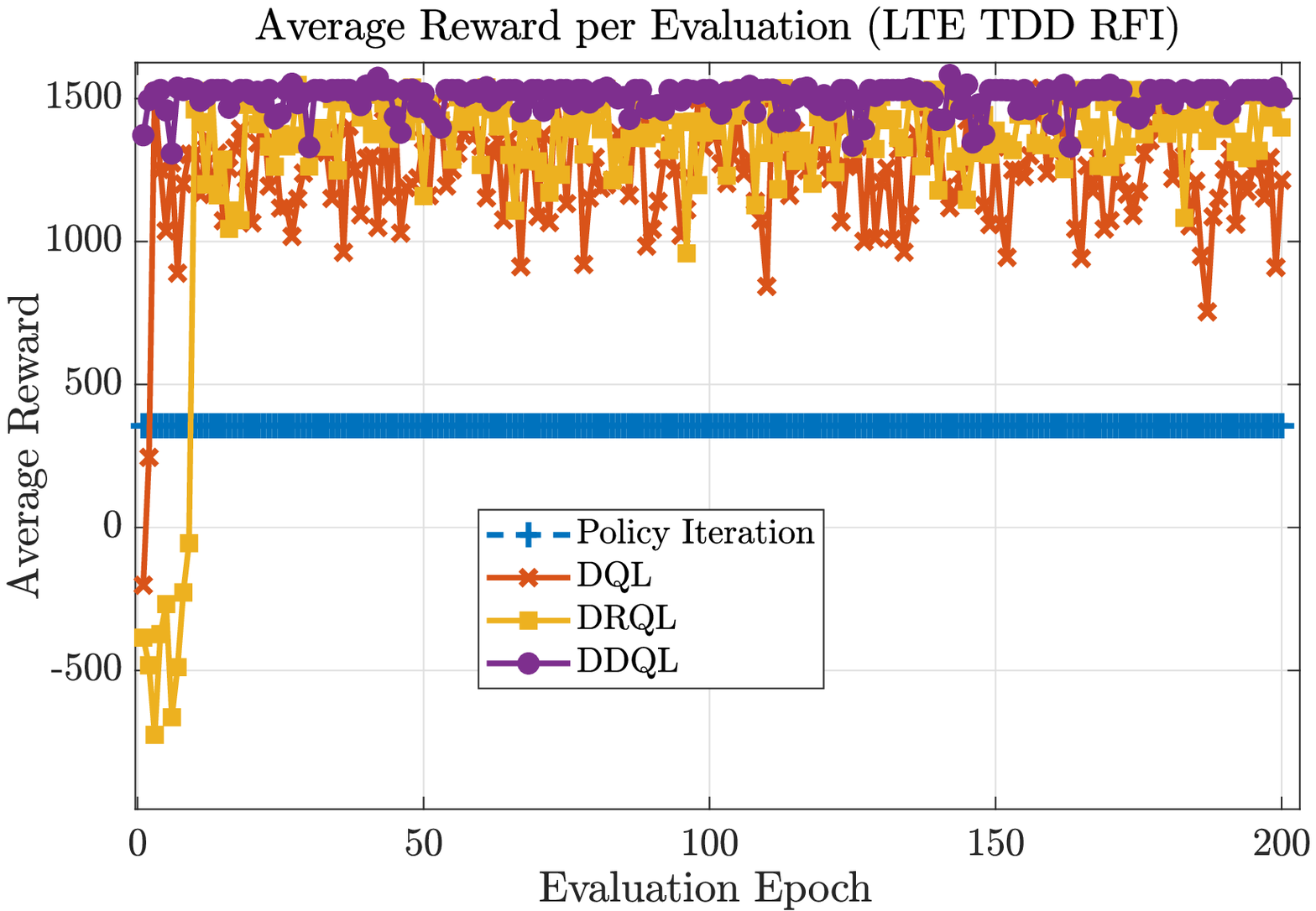}
	\includegraphics[scale=0.5]{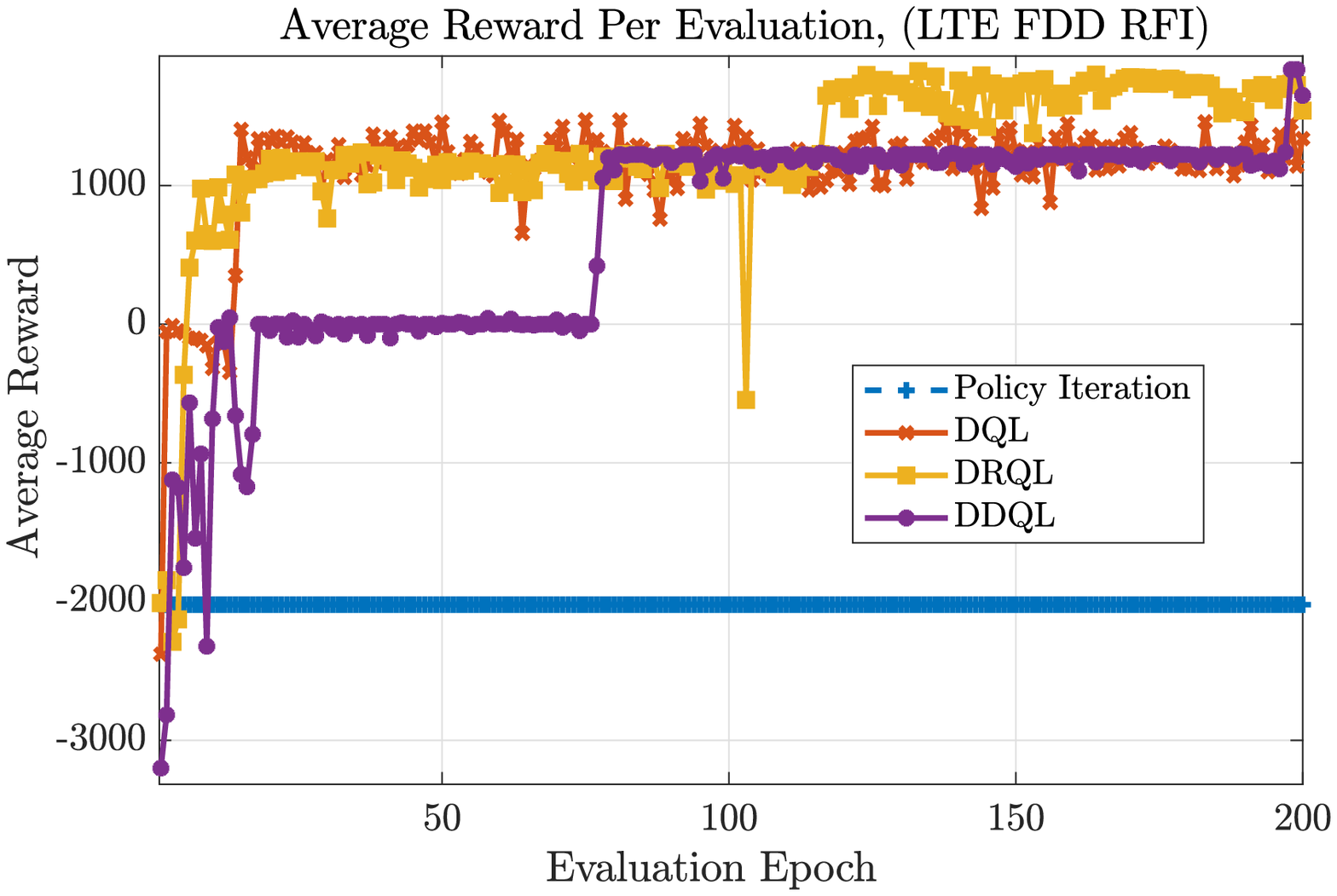}
	\caption{LEFT: Average reward for each RL technique during online evaluation following 200 training epochs in the presence of a LTE TDD communications signal. RIGHT: Average reward during each online epoch in the presence of a LTE FDD communication signal.}
	\label{fig:convTDD}
\end{figure*}

\indent Another DQL extension proposed is DRQL. This extension uses a recurrent Long Short-Term Memory (LSTM) to improve the DQL algorithm's performance when evaluated on MDPs that are not fully-observable \cite{drqn}. A Partially Observable MDP (POMDP) is any MDP where the underlying state can not be directly observed, often due to limited availability of information or environmental noise. Since many real-world problems are partially observable, the Markovian property often doesn't hold in practice. DRQL attempts to mitigate the impact of partial observability with the addition of an LSTM unit, which uses feedback gates that can be used to remember values over arbitrary time intervals. LSTM can be used to resolve long-term dependencies by integrating information across frames. In order to capture time-dependence, DRQL performs SGD updates on samples of \textit{episodes} of transitions, $Ep = \{\phi_{1},\phi_{2},..,,\phi_{q}\},$ as opposed to individual transitions $\phi(s,a,r,s')$. Episodes are sampled randomly from experience memory and updates proceed forward in time through the entire episode. For both DDQL and DRQL, the offline and online learning phases remain the same as in the case of DQL.

\indent Now that we have discussed MDPs, policy iteration, and several Deep RL techniques, we proceed to an experimental comparison of these techniques for control of a cognitive radar operating in a congested spectral setting.

\section{Experimental Methods}
To validate the utility of the RL techniques discussed above for cognitive radar applications, we present experimental results from a COTS implemented cognitive radar prototype. In these experiments, the radar operates in 100MHz of spectrum that must be shared with a communications system. We assume the communications system is non-cooperative and may also be frequency agile. Our COTS system consists of a USRP X310, which is controlled by a host PC and operates in a low-noise closed-loop system where interference is added via an Arbitrary Waveform Generator (AWG). The radar begins operation with a passive sensing period, where the interference and noise power in each of the five sub-bands is used to compile the state vector $s_{t}$ at sensing time $t$. After the sensing period, the radar begins an offline training phase during which the observed state vectors are used to perform the RL offline learning process on the host PC. Once the offline learning phase has concluded, we begin an evaluation phase where the host PC selects appropriate LFM chirp waveforms to send based on the behavior learned during offline learning. If the approach used is a Deep RL technique, then every $N_{online}$ states the radar updates its beliefs based on rewards obtained from analyzing the actions taken by the radar. To validate our system, we first demonstrate that the radar is able to learn a reward function that we believe will maximize radar performance. We then validate the merit of the RL process by examining the radar's performance in terms of target detection and spectrum sharing capabilities in the presence of LTE interference. In these experiments, we analyze both Time Division Duplexing (TDD) and Frequency Division Duplexing (FDD) coexistence scenarios. 

\section{Experimental Results and Discussion} 

\indent First, we must define a reward function for the radar to learn. As discussed in Section II, the radar's actions must balance two priorities: avoiding interference with the communications system and utilizing the maximum available bandwidth. Additionally, we seek to minimize intra-CPI adaptations so that received radar pulses can be coherently processed. Thus, we use the following reward function $R_{t}$
\begin{equation}
R_{t} = R^{+}_{t} + R^{*}_{t},
\label{eq:rwd}
\end{equation}
where
\begin{equation}
  R^{+}_{t} =
\left \{
\begin{array}{ll}
 -45(N_{c}) & \hspace{0.75cm} if \; \;\ N_{c} \; \geq \; 1 \\
 10(N_{SB}-1) & \hspace{0.75cm} if \;\; N{c} \; = \; 0 \\

\end{array}
\right \},
\end{equation}
and
\begin{equation}
 R^{*}_{t} =
\left\{
 \begin{array}{ll}
 0 & \hspace{1.95cm} if \; \;\ N_{WA} \; \; < \; 20 \\
-20 & \hspace{1.94cm} if \;\; \; N_{WA} \; \; \geq \; 20 \\
\end{array}
\right\},
\end{equation}

\noindent where $N_{c}$ is the number of collisions, or instances where the radar and communications system occupy the same sub-band, $N_{SB}$ is the number of sub-bands used by the radar, and $N_{WA}$ is the number of times the radar changes its waveform within a CPI of 1000 radar pulses. In the following results, the radar's performance will be evaluated in the presence of both recorded LTE TDD and FDD interference signals. The TDD interference operates in the second and third sub-bands, and different waveforms are utilized for the offline training and online evaluation phases respectively. The FDD interference operates in the second and third sub-bands during training, and the evaluation waveform operates in the first and second sub-bands.

\subsection{Convergence of Learning}

\begin{figure*}[t]
	\includegraphics[scale=0.38]{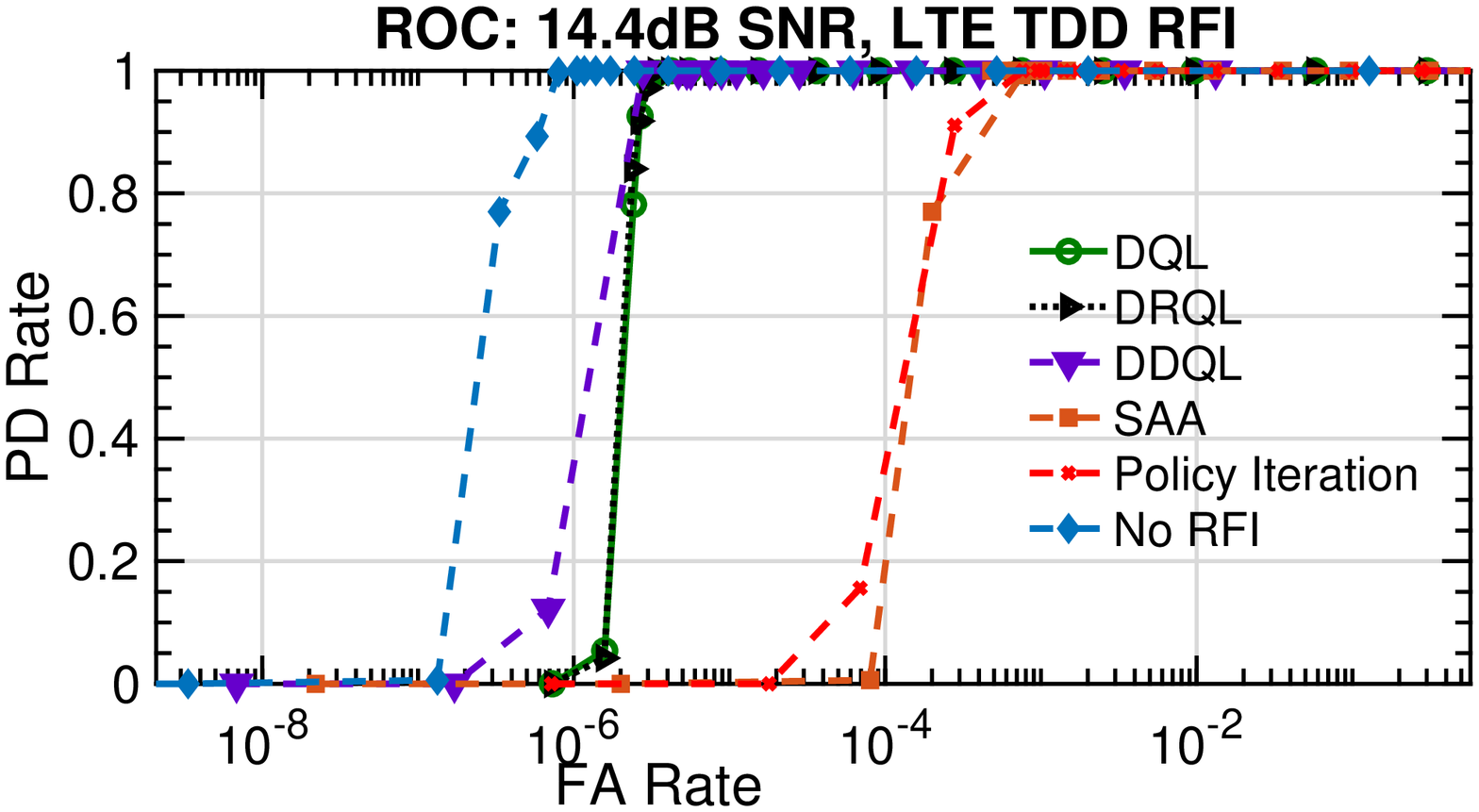}
	\hspace{1.0cm}
	\includegraphics[scale=0.41]{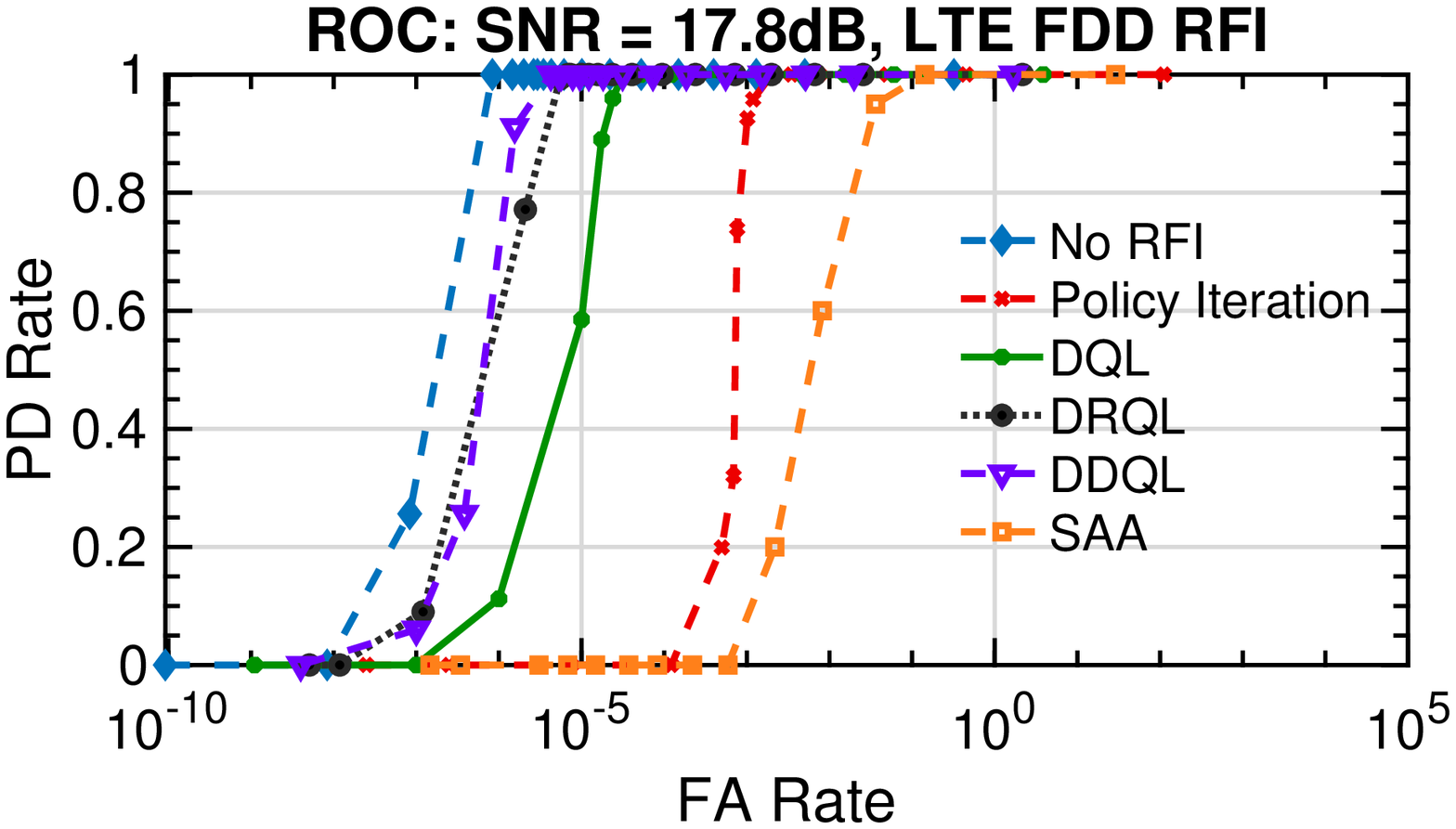}
	\caption{Reciever operating characteristic (ROC) curves for the reinforcement learning COTS radar prototype. Each algorithm attempts to maximize co-channel performance with LTE TDD (LEFT) and FDD (RIGHT) interferers.}
	\label{fig:detection}
\end{figure*}

\indent To determine whether the radar RL techniques can learn desirable behavior from this reward function, the radar is trained and evaluated in an environment with recorded LTE interference. Figure \ref{fig:convTDD} shows the average reward of each RL algorithm during each epoch of evaluation in the presence of both TDD and FDD interference. The radar agents are trained for 200 epochs and evaluated for 200 additional epochs.

\indent In the case of TDD RFI, we see that the policy iteration algorithm converges to a stable solution after the training period, but does not provide as high of an average reward as the Deep RL agents after the off-policy Deep RL agents perform online learning. This is to be expected, since the policy iteration technique only learns during the offline training phase. However, with the exception of DDQL, the Deep RL techniques also do not generalize to the test data immediately and require a period of online learning to converge to a stable solution for the new evaluation environment. Once the Deep RL agents have converged, they are also subject to variability in performance. DQL demonstrates the highest degree of variability for the TDD interference, followed by DRQL, and finally DDQL where the performance remains relatively stable for the entire evaluation phase. Thus, we see that DDQL generalizes very well for this data and also provides a fairly stable solution upon convergence, indicating it has learned an effective strategy for achieving a high reward on average. However, if MDP-PI were trained on environment that is very similar to the evaluation data, it may scale better to the new environment while exhibiting no variability. However, since the evaluation waveform has a different temporal transmission scheme in this case, the lack of online training prevents the MDP from generalizing easily to new environments without additional training.

\indent For the case of FDD RFI all techniques struggle to generalize to the evaluation waveform initially, since the frequency location has shifted as well as the transmission pattern. However, the Deep RL techniques once again achieve fairly stable solutions after the online learning period. Upon convergence, DRQL achieves a slighly higher average reward than DQL indicating that DRQL was able to learn the temporal dependencies slightly better, although both agents are subject to some variability. Although taking a few epochs to converge, DDQL reaches a stable solution with very little variability. After some time, the agent learns it can take a more favorable action and the average reward increases almost immediately, evidenced by the sudden jump at around 75 epochs. Towards the end of evaluation it appears that the agent learns an even more favorable action as the last three epochs show a higher average reward than the second stable solution. While DDQL takes the longest of all the algorithms to converge, it eventually reaches the best solution once again and demonstrates very little variability.

\subsection{Radar Detection Performance}

\indent While these RL techniques have demonstrated utility in maximizing the reward function given in (\ref{eq:rwd}), we must confirm that learning this reward mapping translates to improved radar performance in congested spectrum. We first examine the effect of the RL process on the radar's target detection characteristics. 

\indent Figure \ref{fig:detection} shows Receiver Operating Characteristic (ROC) curves for COTS radar operation guided by each RL technique while an LTE TDD system operates in-band. The radar is training on a downlink reference measurement channel waveform and evaluated on a different TDD downlink waveform. The radar uses a PRI of $409.6 \mu S$ and a CPI consists of 1000 pulses. Each ROC curve corresponds to 500 CPIs of radar data, where each point consists of a different theoretical probability of false alarm $P_{fa}$. The theoretical probability of false alarm is used to analyze detection performance with the Cell Averaging Constant False Alarm Rate (CA-CFAR) algorithm. For every theoretical value of $P_{fa}$, the actual rate of false alarm and missed detections are calculated. These rates are computed from the following equations:
\begin{equation}
\textstyle PD \; \; Rate \; = 1 - \frac{1}{N} \textstyle \sum_{j = 1}^{N} \; MD_{j},
\end{equation}
and
\begin{equation}
\textstyle FA \; \; Rate \; = \textstyle \frac{1}{N \times N_{p}}  \textstyle \sum_{j = 1}^{N} \; FA_{j},
\end{equation}

\noindent where $N$ is the number of CPI's, $N_{p}$ is the number of points in the 2D range-Doppler map, $MD_{j}$ is the observed number of missed detections in CPI $j$, and $FA_{j}$ is the observed number of false alarms in CPI $j$.

\indent In Figure \ref{fig:detection}, we see the detection performance for the radar operation in TDD and FDD interference scenarios, where each RL technique is utilized as well as SAA. In the TDD interference scenario, SAA and policy iteration perform similarly. This is due to the large number of waveform adaptations both approaches utilize, as well as the lack of generalization inherent to the on-policy technique. The Deep RL techniques perform much better in the TDD scenario, noted by the shift of the ROC curves to the left, denoting a higher $PD$ rate for a given $FA$ rate. Among the Deep RL techniques, DDQL comes closest to the no RFI bound, but only provides a slight advantage over DQL and DRQL, which perform similarly in this scenario.

\indent In the FDD scenario, we see that the policy iteration technique performs notably better than SAA, which can be attributed to the high number of waveform adaptations utilized by SAA. Once again, we see a significant improvement from the Deep RL techniques, which begin to approach the best case performance of no RFI. In the FDD scenario, DRQL performs slightly better than DQL and very similar to DDQL. This is in line with what we would expect, as Figure \ref{fig:convTDD} showed that DRQL receives a slightly higher average reward than DQL after the online training period and a similar average reward to DDQL.

\subsection{Spectrum Sharing Performance}

We also evaluate the spectrum sharing capabilities of each approach in the presence of LTE TDD and FDD interferers. Coexistence capabilities will be examined in terms of the number of collisions, number of missed opportunities, and percentage of total transmissions where waveform adaptation occurs. Collisions correspond to the number of time-frequency slots where the radar and communication system are transmitting in the same frequency band concurrently. The number of missed opportunities corresponds to the number of open time-frequency slots that were available to the radar but not utilized.

In the case of TDD interference, the policy iteration approach collides occasionally with the RFI. This is because of the inherent difference between the training and test data. However, the radar is able to avoid more of the RFI transmits than the SAA technique with a similar number of waveform adaptations. The Deep RL approaches fair much better in this case, avoiding almost all the interference with a low number of waveform adaptations. However, the RL agents utilize less available spectrum than the SAA technique. However, the decreased number of collisions and less waveform adaptations lead to better detection characteristics than SAA, indicating better performance from both a radar and communications perspective.

In the case of FDD RFI, the policy iteration technique performs very poorly due to the new frequency location of the interferer upon evaluation. The policy iteration agent has not seen the evaluation interference state in training and will default to transmitting only in the first band, where the interference also lies. The Deep RL techniques once again result in very few collisions with a slightly higher number of missed opportunities than the SAA case. Among the Deep RL techniques, DRQN and DDQN result in similar performance to DQN with less waveform adaptations. 

\begin{table}[t]
	\centering
	\caption{Comparison of Spectrum Sharing Performance in an LTE TDD RFI Environment (TOP) and LTE FDD RFI Environment (BOTTOM).}
	\label{tab:my-table1}
	\begin{tabular}{|l|l|l|l|}
		\hline		
	    Technique & \# Collisions & \# Missed Opp. & \% Waveform Adapt. \\ \hline
		Policy Iteration & 21 & 120 & 42.57\% \\ \hline
		DQN & 5 & 119 & 11.89\% \\ \hline
		DRQN & 5 & 114 & 2.97\% \\ \hline
		DDQN & 0 & 113 & 5.94\% \\ \hline
		SAA  & 44 & 66 & 43.56\% \\ \hline
		 
	\end{tabular}
\end{table}

\begin{table}[t]
	\centering
	\label{tab:my-table2}
	\begin{tabular}{|l|l|l|l|}
		\hline
	    Technique & \# Collisions & \# Missed Opp. & \% Waveform Adapt. \\ \hline
		Policy Iteration & 63 & 260 & 0\% \\ \hline
		DQN & 2 & 69 & 15.84\% \\ \hline
		DRQN & 2 & 60 & 3.96\% \\ \hline
		DDQN & 1 & 58 & 1.98\% \\ \hline
		SAA  & 30 & 36 & 54.46\% \\ \hline
		
	\end{tabular}
\end{table}

\section{Conclusion}

Here, we have discussed the principles of several state-of-the-art RL algorithms and applied each technique to cognitive radar waveform selection on a hardware-implemented prototype. From our experimental results, we conclude that an RL approach for waveform selection leads to improved target detection characteristics in a congested spectral environment relative to SAA. Additionally, the RL approaches result in decreased mutual interference with a communications system relative to SAA at the cost of slightly lower bandwidth utilization. However, based on the improved detection performance, the lower bandwidth utilization does not seem to be a major issue for this application. Among the RL algorithms, we note that policy iteration is useful for its stability upon convergence as well as lower inherent estimation error as the technique directly solves the MDP we propose. However, once a policy has been established, real-time online learning is impractical. Additionally, the algorithm involves iteratively solving a set of equations that require large transition probability and reward matrices, and the computational complexity thus grows exponentially with the state space size. 

\indent The off-policy Deep RL techniques are a useful alternative to policy iteration as the solution does not depend on explicitly modeling the transition probability function. Additionally, since DQL is an off-policy approach, the radar can continue updating its beliefs while transmitting. However, training the neural networks required for this approach is resource intensive and some instability is apparent even after a training period. Among the Deep RL algorithms examined here, we note that DRQL converges to a slightly more favorable solution in terms of our reward function that DQL, provided that there is some temporal dependency in the data, which is true of the LTE interference analyzed here. Additionally, DDQL results in a very stable learning process that results in the most favorable solutions in the LTE TDD and FDD RFI cases shown here. Thus, we can conclude that the DQL and DRQL techniques result in some optimization errors that are avoided by the additional validation step in DDQL. 

\indent The experimental analysis presented here demonstrates that RL techniques are viable strategies for control of a spectrum sharing radar, given that the state space of the environmental model remains tractable and sufficient offline and online training time is allotted. RL allows for long-term planning, which can be used to develop efficient action patterns that result in little target distortion compared to SAA. However, it is possible that other environmental models may be more appropriate for a true spectrum sharing environment. Future work will focus on modeling a realistic coexistence environment with many users and guiding the RL process with domain knowledge.

\end{document}